\documentclass[letterpaper]{article} 
\usepackage{aaai25}
\usepackage{times}  
\usepackage{helvet}  
\usepackage{courier}  
\usepackage[hyphens]{url}  
\usepackage{graphicx} 
\usepackage{natbib}  
\usepackage{caption} 
\usepackage{algorithm}
\usepackage{algorithmic}
\usepackage{arydshln}
\usepackage{newfloat}
\usepackage{listings}
\DeclareCaptionStyle{ruled}{labelfont=normalfont,labelsep=colon,strut=off} 
\floatstyle{ruled}
\newfloat{listing}{tb}{lst}{}
\floatname{listing}{Listing}
\title{CollabEval: Enhancing LLM-as-a-Judge via  Multi-Agent Collaboration}
\author{
    Yiyue Qian \textsuperscript{\rm 1}\equalcontrib, Shinan Zhang  \textsuperscript{\rm 1}\equalcontrib, Yun Zhou \textsuperscript{\rm 2}, Haibo Ding \textsuperscript{\rm 2}, Diego Socolinsky \textsuperscript{\rm 1}, Yi Zhang \textsuperscript{\rm 2}
 \\
}
\affiliations{
     \{iamyiyue, shinanz, yunzzhou, hbding, sclinsky,  yizhngn\}@amazon.com \\
    \textsuperscript{\rm 1}Amazon AWS Generative AI Innovation Center\\
    \textsuperscript{\rm 2}Amazon AWS Bedrock\\

}
\usepackage{bibentry}

\usepackage[utf8]{inputenc} 
\usepackage[T1]{fontenc}    
\usepackage{url}            
\usepackage{booktabs}       
\usepackage{amsfonts}       
\usepackage{nicefrac}       
\usepackage{caption}
\usepackage{microtype}      
\usepackage{algorithmic}
\usepackage{algorithm}
\usepackage{enumitem}
\usepackage{graphicx}
\usepackage{multirow}
\usepackage{amsthm}
\usepackage{amsmath}
\usepackage{amssymb}
\theoremstyle{definition}

\theoremstyle{plain}

\begin{document}
\maketitle
\begin{abstract}
Large Language Models (LLMs) have revolutionized AI-generated content evaluation, with the LLM-as-a-Judge paradigm becoming increasingly popular. However, current single-LLM evaluation approaches face significant challenges, including inconsistent judgments and inherent biases from pre-training data. To address these limitations, we propose \textbf{CollabEval}, a novel multi-agent evaluation framework that implements a three-phase \textbf{Collab}orative \textbf{Eval}uation process: initial evaluation, multi-round discussion, and final judgment. Unlike existing approaches that rely on competitive debate or single-model evaluation, CollabEval emphasizes  collaboration among multiple agents with strategic consensus checking for efficiency. Our extensive experiments demonstrate that CollabEval consistently outperforms single-LLM approaches across multiple dimensions while maintaining robust performance even when individual models struggle. The framework provides comprehensive support for various evaluation criteria while ensuring efficiency through its collaborative design. 
\end{abstract}
%
\section{Introduction}
The rapid advancement of Large Language Models (LLMs) has revolutionized AI-generated content evaluation, making the LLM-as-a-Judge paradigm increasingly popular~\citep{chiangchatbot,wang2024self,raina2024llm,chanchateval,li2025verification,maautodata,song2025learning,ma2025llm,qian2025enhancing}. Recent studies have demonstrated the potential of using single LLMs as evaluators, with notable work~\cite{bai2024mt} introducing MT-bench for systematic LLM evaluation, and another work ~\cite{chiangchatbot} developing Chatbot Arena as an open platform for LLM assessment through human preference alignment. These approaches have shown promising results in automating evaluation processes across various dimensions including coherence, relevance, and fluency.  However, recent studies have identified significant limitations in current evaluation methodologies. One recent research  ~\cite{raina2024llm} reveals that LLM-based evaluations are vulnerable to universal adversarial attacks, raising concerns about their reliability. Additionally, Wang et al. ~\cite{wang2024self} demonstrated that self-taught evaluators often struggle with consistency and objectivity in their assessments, highlighting the critical need for more robust evaluation frameworks.

Generally speaking, current evaluation methodologies face several critical challenges: (i). single-LLM evaluations lack robustness due to inherent biases from their pre-training data and knowledge~\cite{huang2024c}. Recent studies~\cite{bai2024mt,huang2024c} have found significant variations in judgment quality across different LLM evaluators, with ChatEval~\cite{chanchateval} further highlighting that individual LLMs may excel in certain dimensions while underperforming in others.  (ii). While recent works~\cite{chanchateval,chen2023reconcile,chern2024can} have developed agent-based frameworks to address these limitations, with ChatEval~\cite{chanchateval} notably implementing multiple debate agents for evaluation, these approaches often lack the flexibility and efficiency needed for diverse evaluation scenarios. These challenges underscore the need for a more robust and adaptable evaluation framework.

To address these limitations, we present \textbf{CollabEval}, a novel multi-agent evaluation framework that implements a structured (i.e., three-phrase) collaborative assessment process. Unlike previous approaches~\cite{chanchateval,chen2023reconcile}, our framework emphasizes collaboration rather than competitive debate, addressing the need for diverse model perspectives in evaluation as identified by~\cite{verga2024replacing}. Specifically, CollabEval employs a sophisticated three-phase design: (1) initial evaluation, where different agents independently assess the content; (2) multi-round collaborative discussion, where agents share and refine their evaluations through structured dialogue, including confidence scores, agreements, disagreements, and reasoning; and (3) final judgment, where ultimate evaluation decisions are made based on prior discussions. Notably, CollabEval performs consensus checks at each phase, allowing for early termination when agreement is reached, thus ensuring efficiency compared to existing agent-based LLM-as-a-Judge methods. The key contributions of our work include: \begin{itemize}[leftmargin=0.1in]\setlength{\itemsep}{0.2pt}
\item \textbf{Novel}: We introduce a three-stage evaluation framework that uniquely combines independent assessment with collaborative refinement among agents. 
\item \textbf{Comprehensive}: CollabEval supports both criteria-based and pairwise comparisons across multiple dimensions, demonstrating superior performance over single-LLM evaluations via extensive experimental validation. 
\item \textbf{Robust and Efficient}: Our framework maintains strong performance even when individual LLMs show weaknesses, while ensuring efficiency through strategic consensus checking and early termination.
\end{itemize}

\begin{figure*}[t]
 	\centering
 	\includegraphics[width=1.0\textwidth]{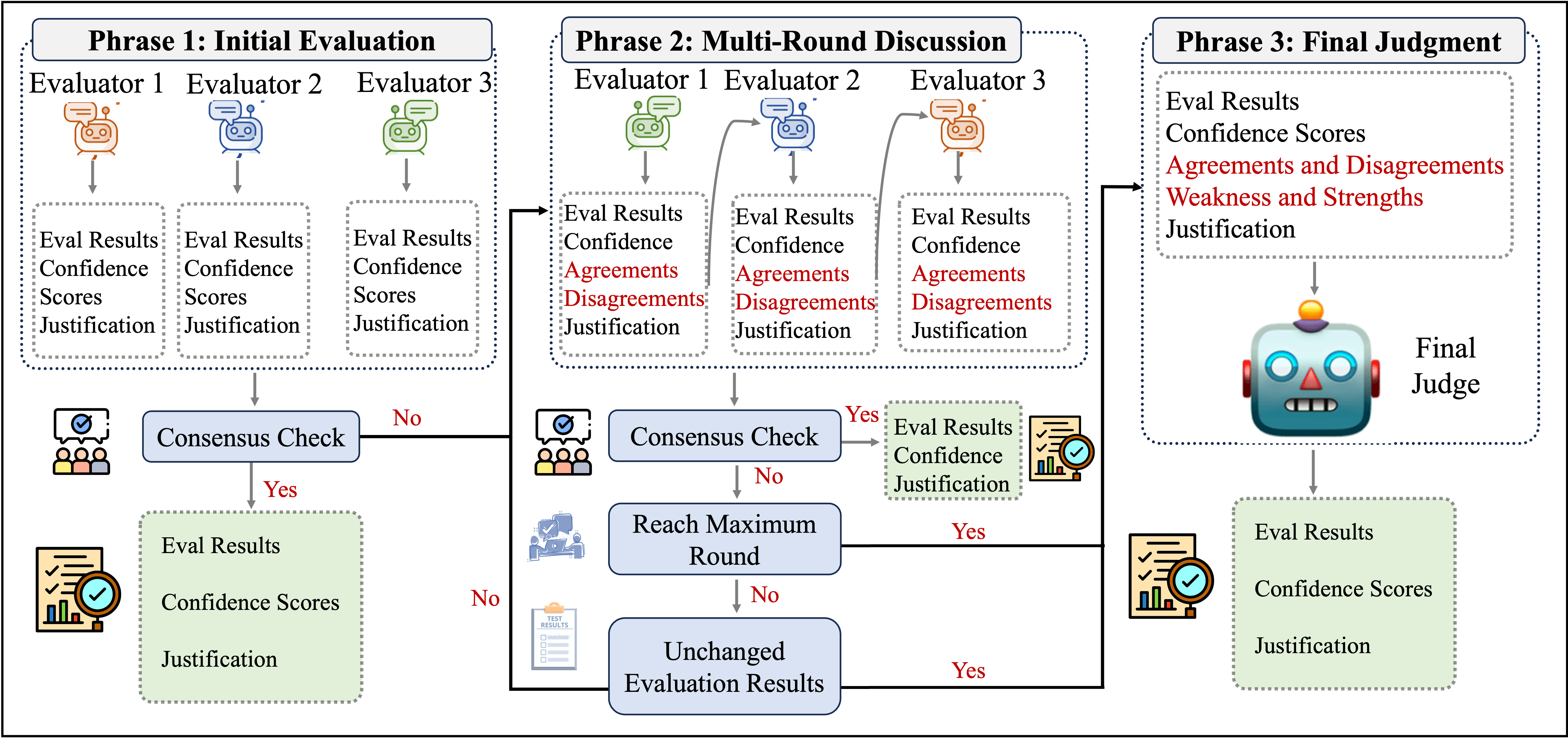}
         	\caption{The framework of CollabEval consists of three main phases: (a) Phase 1: Initial Evaluation - Three evaluators independently assess content, providing evaluation results, confidence scores, and justifications. A consensus check is performed; if consensus is reached, final results are returned, otherwise proceeding to Phase 2. (b) Phase 2: Multi-Round Discussion - Evaluators engage in collaborative discussion sharing agreements, disagreements, and justifications. After each round, a consensus check is performed. If consensus is reached, results are returned; if not, the system checks for maximum rounds or unchanged results before proceeding. (c) Phase 3: Final Judgment - When consensus cannot be reached through discussion, a final judge analyzes all previous evaluation results, confidence scores, agreements/disagreements, and justifications to make the ultimate evaluation decision.
        } 
 	\label{structure}
 \end{figure*}
\section{Related Work}
\noindent\textbf{LLM-as-a-Judge.}
Recent advances in LLMs have led to increasing adoption of the LLM-as-a-Judge paradigm for evaluating AI-generated content. Bai et al. introduced MT-bench as a systematic framework for LLM evaluation, establishing benchmarks for assessing model performance across various dimensions~\cite{bai2024mt}. Chiang et al. developed Chatbot Arena as an open platform leveraging human preference alignment for LLM assessment, demonstrating the potential of structured evaluation frameworks~\cite{chiangchatbot}. However, existing single-LLM approaches face significant limitations. Raina et al. revealed critical vulnerabilities to universal adversarial attacks in LLM-based evaluations~\cite{raina2024llm}, while Wang et al. demonstrated that self-taught evaluators struggle with consistency and objectivity~\cite{wang2024self}. Huang et al. further highlighted how single-LLM evaluations often lack robustness due to inherent biases from their pre-training data and knowledge~\cite{huang2024c}, showing significant variations in judgment quality across different LLM evaluators.
\\
\\
\noindent\textbf{Multi-agents in LLMs.} 
Recent research has explored multi-agent approaches~\cite{chen2023reconcile,hongmetagpt,shah2024multi,wang2024rethinking,wuautogen,zhang2024towards,han2024llm} for enhancing LLM capabilities across various tasks. For instance, ReConcile~\cite{chen2023reconcile}, a framework that improves reasoning through round-table conferences among diverse LLMs. Their approach enables collaborative reasoning between LLM agents via multiple rounds of discussion, incorporating confidence-weighted voting mechanisms to achieve better consensus. 

In the context of LLM-as-a-Judge, several works~\cite{zhuge2024agent,chanchateval,chern2024can,rasheed2024large} have explored multi-agent evaluation frameworks. ChatEval~\cite{chanchateval} is developed by implementing multiple debate agents for autonomous discussion and evaluation of AI-generated content. It showed that collaborative evaluation through debate can lead to more reliable assessments. Besides, Chern et al. investigated the potential of agent debate for meta-evaluation~\cite{chern2024can}. These approaches demonstrated that multi-agent evaluation systems can effectively address the limitations of single-LLM judges, particularly in terms of robustness and consistency. However, many existing approaches rely heavily on competitive debate rather than collaboration, potentially leading to inefficiencies in the evaluation process. This limitation motivates our work on CollabEval, which emphasizes collaboration over competitive debate to achieve more reliable and efficient evaluations.
\section{Proposed Framework}
In this section, we present the details of CollabEval  including three main phrases: initial evaluation, multi-round collaborative discussion among agents, and final judgement .~\label{methodology}
\subsection{Initial Evaluation}
Single LLM evaluators often exhibit inherent biases stemming from their pre-training data and knowledge bases. These biases, coupled with varying pre-training datasets and knowledge across different LLMs, can lead to inconsistent judgments when evaluating AI-generated content. To address these limitations and leverage the diversity of different LLMs, we propose a multi-agent collaborative evaluation framework.

In Phase 1, as illustrated in Figure~\ref{structure}, CollabEval employs multiple independent evaluators to conduct initial assessments. Each evaluator independently analyzes the content and provides three key components: evaluation results, confidence scores, and detailed justifications for their assessments. This independent evaluation ensures that each agent's unique perspective and capabilities are captured without influence from other evaluators. Once all evaluators complete their initial assessments, CollabEval performs a consensus check to determine whether the evaluators have reached agreement in their judgments. If consensus is achieved, the system returns the final evaluation results, demonstrating efficient early termination. However, if evaluators show significant disagreement, the process advances to Phase 2, where evaluators engage in multi-round discussions to resolve differences and refine their assessments.

\subsection{Multi-Round Discussion}

\textbf{Agents Collaboration.}
During  Phase 2, evaluators share their initial evaluations, confidence scores, and justifications with each other. The collaboration focuses on identifying agreements and disagreements in their assessments. Each evaluator reviews others' evaluations and provides updated assessments based on the collective insights. This process enables evaluators to refine their judgments by incorporating multiple perspectives and addressing potential biases or oversights in their initial evaluations.
\\  
\\
\noindent\textbf{Iterative Process.}
The discussion proceeds iteratively, with evaluators using all available data from both initial evaluations and ongoing discussions to refine their assessments. Each evaluator considers:
\begin{itemize}[leftmargin=0.3in]\setlength{\itemsep}{0.2pt}
\item Initial evaluation results from all agents
\item Confidence scores from previous rounds
\item Areas of agreement and disagreement from  other evaluators
\item Justifications provided by other evaluators
\end{itemize}

For instance, as illustrated in Figure~\ref{structure}, at the 1-round discussion, Evaluator 1 begins by analyzing all initial evaluation results and provides updated assessments with specific agreements and disagreements. Evaluator 2 then considers both the initial evaluations and Evaluator 1's updated assessment before providing its refined evaluation. Finally, Evaluator 3 reviews all previous evaluations - both initial and updated - before contributing its assessment. To mitigate potential biases from model capabilities, we randomly shuffle the order of evaluators in discussion rounds. 
\\
\\
\noindent\textbf{Consensus Check.}
After each discussion round, CollabEval performs three critical checks to determine the next steps in the evaluation process. First, the system examines whether all evaluators have reached consensus on their evaluations at the current-round discussion. If consensus is achieved, the system returns the final results. Otherwise, CollabEval then proceeds to verify two additional conditions: whether the maximum number of discussion rounds has been reached, and whether the evaluation results remain unchanged from the previous round. When any of these two conditions are met - either the maximum rounds are reached, or evaluations remain static - the process advances to the final evaluation stage. However, if none of these conditions are satisfied, CollabEval initiates another round of discussion to further refine the evaluations.

\subsection{Final Judge Evaluation}
When the multi-round discussion fails to reach consensus or evaluations remain unchanged, CollabEval employs a strong model as the final judge. The final judge makes the ultimate evaluation decision by analyzing all evaluation results from previous rounds, confidence scores and justifications, areas of agreement and disagreement among evaluators, and the progression of evaluations through discussion rounds. The final judge considers this comprehensive information to provide a decisive assessment that considers all perspectives and reasoning presented during the evaluation process.


\section{Experiments} 
In this section, we present a comprehensive evaluation of CollabEval across two distinct evaluation modes: criteria-based evaluation and pair-wise comparison. We conduct experiments using three benchmark datasets to assess the framework's performance. Finally, we discuss key findings and insights derived from our experiments.

\subsection{Experiment Setup}
\subsubsection{Evaluation Mode.}
To comprehensively evaluate the capability of our CollabEval, we conduct experiments in two distinct evaluation modes:
\\
\noindent\textit{Criteria-based Evaluation:} This mode assesses content across multiple pre-defined dimensions, such as coherence, consistency, fluency, and relevance. Each dimension is scored on specific scales, allowing for fine-grained assessment of different aspects of the generated content.
\\
\noindent\textit{Pair-wise Comparison:} 
In this mode, evaluators determine which of two responses is better by directly comparing them. This approach is particularly useful for relative quality assessment and helps establish preference rankings between different model outputs.

\subsubsection{Datasets.}
We utilize three benchmark datasets including one criteria-based dataset (i.e., SummEval dataset~\cite{fabbri2021summeval}) and two pair-wise comparison dataset (i.e., chatbot\_arena\_conversation dataset~\cite{chatbotarena2023} and lmsys\_arena\_human\_preference\_55k dataset~\cite{chiangchatbot}). Next, we will introduce more details about these benchmark datasets.
\\
\noindent\textit{Criteria-based Dataset:}
We first utilize SummEval~\cite{fabbri2021summeval}, a comprehensive benchmark dataset containing 1600 examples generated from 100 source news articles. These summaries are produced by 16 different models, ensuring a diverse range of generation qualities and styles. Each summary undergoes rigorous evaluation by 8 expert annotators across four critical dimensions: coherence, consistency, fluency, and relevance. The scoring system employs a 5-point scale ranging from 1 to 5, allowing for fine-grained assessment of quality. The dataset is structured in a detailed format including (id, machine\_summary, source\_news, coherence\_score, consistency\_score, fluency\_score, relevance\_score), enabling comprehensive analysis of each dimension independently.
\\
\noindent\textit{Pair-wise Comparison Dataset:}
For pair-wise comparison evaluation, we employ two distinct datasets. Specifically, for the chatbot\_arena\_conversations ~\cite{chatbotarena2023} dataset, instead of using all datasets, we randomly select 1,000 samples. This dataset focuses on direct comparisons between different model responses in conversational settings. Besides, for the lmsys\_arena\_human\_preference\_55k dataset~\cite{chiangchatbot}, we also utilize 1,000 random samples. This dataset is particularly valuable as it incorporates human preference judgments, providing a robust ground truth for evaluation.
Both datasets follow a standardized format of (id, query, response\_a, response\_b, winner), enabling direct comparison between two alternative responses and clear identification of the superior option.

\subsubsection{Baseline.}
In this work, we compare our model CollabEval with two groups of baseline methods including single LLM-as-a-Judge and Agent-based LLM-as-a-Judge. 
\\
\noindent\textit{\textbf{B1}: Single LLM-as-a-Judge:} 
This baseline represents the traditional approach where a single LLM evaluates content independently. We implement multiple state-of-the-art models including Mistral Large~\cite{mistral2024large}, Claude Haiku~\cite{anthropic2024claude3}, Claude Sonnet 3 ~\cite{anthropic2024claude3}, and Llama 3 70b~\cite{meta2024llama3} as individual evaluators. Each model serves as an independent judge to demonstrate the capability of relying on individual model judgments and to establish a performance benchmark for comparison.
\\
\noindent\textit{\textbf{B2}: Agent-based LLM-as-a-Judge:} 
For agent-based approach, we also explored another round-table discussion mechanism, called Round-Table Agents Eval in Table~\ref{tab:criteria}. Instead of following the three-stage mechanism, we follow the round-table mechanism in this work~\cite{chen2023reconcile} and implement a sequential round-table discussion where agents evaluate content one after another. Specifically, for each evaluation task, we randomly select one agent to provide an initial assessment. The next agent then reviews this evaluation, provides its own assessment, and either agrees with or revises the previous evaluation. This process continues sequentially through all agents. To reduce potential biases from agent ordering, we randomly shuffle the sequence of agents for each new evaluation task. The discussion continues until either all agents reach consensus or a maximum of rounds is completed. If no consensus is reached after the maximum rounds, a majority voting mechanism is applied to determine the final evaluation result. This more sophisticated baseline implements a round-table discussion approach where multiple LLMs engage in collaborative evaluation. This method serves as an intermediate step between single-agent and our proposed CollabEval approach.

\subsubsection{Experimental Setting.}
CollabEval employs multiple state-of-the-art LLM agents (Mistral Large, Claude Haiku, Claude Sonnet, and Llama 3 70b) for evaluation. In Phase 1, each agent independently provides initial assessments. To mitigate potential biases from model ordering, we randomly shuffle the sequence of evaluators in both the initial evaluation and multi-round discussion phases. During Phase 2, if consensus is not reached initially, the evaluation process continues through multiple rounds of discussion, with a maximum of 3 rounds. If consensus remains unachieved after the discussion phase, we employ Claude Sonnet 3.5~\cite{anthropic2024sonnet} as the final judge in Phase 3, leveraging its strong reasoning capabilities to analyze the comprehensive evaluation history and make the ultimate decision.

\begin{table*}[h]
\caption{Comparison results among CollabEval and single LLM-as-a-Judge on SummEval dataset for criteria-based evaluation. Best accuracy for each dimension is in bold.}
\label{tab:criteria}
\scalebox{0.94}{
\begin{tabular}{lcccccccc}
\toprule
Model & Accuracy & Avg & Gap 1 & Gap 2 & Gap 3 & Gap 4 & Over- & Under- \\
Setting & (\%) & Rounds & Ratio (\%) & Ratio (\%) & Ratio (\%) & Ratio (\%) & eval Ratio (\%)& eval Ratio (\%)\\
\hline
\hline
\multicolumn{9}{l}{\textbf{Relevance}} \\
\hline
CollabEval & \textbf{49.5} & 2.073 & 87.8 & 12.0 & 0.5 & 0 & 31.9 & 68.1 \\
\hdashline
Single-LLM Sonnet & 47.7 & 1 & 85.5 & 13.7 & 1.6 & 0 & 29.7 & 70.3 \\
Single-LLM Haiku & 47.6 & 1 & 84.9 & 14.7 & 1.1 & 0 & 30.2 & 69.8 \\
Single-LLM Llama3 & 22.8 & 1 & 76.7 & 23.3 & 0.0 & 0 & 100.0 & 0.0 \\
\hline
\hline
\multicolumn{9}{l}{\textbf{Coherence}} \\
\hline
CollabEval & \textbf{40.4} & 2.343 & 77.8 & 20.8 & 1.5 & 0 & 63.3 & 36.7 \\
\hdashline
Single-LLM Sonnet & 37.4 & 1 & 71.4 & 23.9 & 4.9 & 0 & 66.4 & 33.6 \\
Single-LLM Haiku & 38.9 & 1 & 76.9 & 22.4 & 0.8 & 0 & 63.4 & 36.6 \\
Single-LLM Llama3 & 29.5 & 1 & 77.0 & 22.0 & 2.2 & 0 & 25.4 & 74.6 \\
\hline
\hline
\multicolumn{9}{l}{\textbf{Fluency}} \\
\hline
CollabEval & \textbf{46.9} & 2.103 & 77.8 & 18.0 & 4.5 & 0 & 21.9 & 78.1 \\
\hdashline
Single-LLM Sonnet & 46.8 & 1 & 65.9 & 24.0 & 21.4 & 5 & 29.7 & 70.3 \\
Single-LLM Haiku & 13.8 & 1 & 75.9 & 22.3 & 6.2 & 0 & 30.2 & 69.8 \\
Single-LLM Mistral & 45.8 & 1 & 86.7 & 13.3 & 0.0 & 0 & 25.0 & 75.0 \\
\hline
\hline
\multicolumn{9}{l}{\textbf{Consistency}} \\
\hline
CollabEval & \textbf{48.2} & 2.181 & 79.6 & 18.2 & 7.0 & 0 & 10.2 & 89.8 \\
\hdashline
Single-LLM Sonnet & 46.9 & 1 & 65.8 & 25.2 & 19.8 & 0 & 10.4 & 89.6 \\
Single-LLM Haiku & 12.6 & 1 & 77.7 & 20.9 & 5.3 & 0 & 4.7 & 95.3 \\
Single-LLM Mistral & 55.9 & 1 & 93.8 & 5.4 & 2.8 & 0 & 24.4 & 75.6 \\
\bottomrule
\end{tabular}}
\end{table*}

\begin{table*}[h]
\caption{Comparison results among multi-agents and single LLM-as-a-Judge on two Arena datasets for pairwise comparison evaluation. Best accuracy for each dataset is in bold.}
\label{tab:pairwise}
\vspace{-0.1in}
\scalebox{0.95}{\begin{tabular}{lcccc}
\toprule
Model Setting & Accuracy (\%)  & Average Rounds & GT\_Win \_Pred\_Tie Ratio(\%) & GT\_Tie\_Pred\_Win Ratio (\%)\\
\hline
\hline
\multicolumn{5}{l}{\textbf{Chatbot Arena Data}} \\
\hline
CollabEval (Ours) & \textbf{60.2} & 1.542 & 50.00 & 2.63 \\
Round-Table Agents Eval & 57.7 & 1.214 & 15.84 & 43.97 \\
\hdashline
Single-LLM Mistral Large & 58.2 & 1 & 45.54 & 4.22 \\
Single-LLM Haiku & 57.2 & 1 & 46.30 & 3.38 \\
Single-LLM Llama3 70b & 59.7 & 1 & 53.85 & 0.00 \\
\hline
\hline
\multicolumn{5}{l}{\textbf{Arena Human Preference Data}} \\
\hline
CollabEval (Ours) & \textbf{51.5} & 1.517 & 53.20 & 9.07 \\
Round-Table Agents Eval & 48.7 & 1.258 & 12.70 & 47.37 \\
\hdashline
Single-LLM Sonnet & 48.4 & 1 & 48.06 & 13.95 \\
Single-LLM Mistral Large & 50.5 & 1 & 54.95 & 5.25 \\
Single-LLM Llama3 70b & 48.8 & 1 & 55.47 & 0.39 \\
\bottomrule
\end{tabular}}
\end{table*}

\subsection{Performance Discussion}
\subsubsection{Discussion about criteria-based evaluation.}
Table~\ref{tab:criteria} shows the comparison results of all methods for criteria-based evaluation on SummEval dataset.  
This table employs several key metrics to assess performance. \textbf{Accuracy} measures the percentage of correct evaluations compared to ground-truth labels. \textbf{Average Rounds} indicates the number of discussion iterations required for evaluators to reach consensus. The \textbf{Gap Ratios} (1-4) measure the  percentage of samples having absolute difference between LLM-assigned scores and ground-truth labels among all misevaluated samples, where Gap 1 represents a one-point difference, Gap 2 a two-point difference, and so on. The evaluation bias is captured through \textbf{Over-evaluation Ratio}, indicating the percentage of misevaluated samples where LLM scores exceed ground-truth labels among all misevaluated samples, and \textbf{Under-evaluation Ratio}, where scores fall below ground-truth labels.

Our experimental results demonstrate CollabEval's superior performance across all evaluation dimensions. In relevance assessment, CollabEval achieves 49.5\% accuracy with 2.073 average rounds, showing the highest Gap 1 Ratio (87.8\%) and minimal severe misjudgments (0.5\% Gap 3, 0\% Gap 4). The coherence evaluation reveals CollabEval's robust performance with 40.4\% accuracy and balanced error distribution (77.8\% Gap 1, 20.8\% Gap 2, 1.5\% Gap 3), significantly outperforming single-LLM approaches. For fluency assessment, CollabEval maintains competitive accuracy (46.9\%) while demonstrating better error distribution (77.8\% Gap 1, 18.0\% Gap 2, 4.5\% Gap 3) compared to Single-LLM Sonnet's more scattered profile. In consistency evaluation, CollabEval achieves 48.2\% accuracy with the most balanced error distribution (79.6\% Gap 1, 18.2\% Gap 2, 7\% Gap 3). 

Notably, while requiring additional computational rounds (average 2.073-2.343), CollabEval consistently shows more balanced over/under-evaluation ratios across all dimensions, particularly evident in relevance (31.9\%/68.1\%) and coherence (63.3\%/36.7\%), demonstrating significant improvement over single-LLM approaches such as Llama3's extreme 100\%/0\% split in relevance evaluation. The analysis of over-evaluation and down-evaluation ratios reveals distinct behavioral patterns across different LLMs in their evaluation tendencies. Most notably, Llama3 exhibits extreme evaluation patterns, showing a 100\% over-evaluation ratio in the relevance dimension, indicating a consistent bias toward higher scores. This contrasts sharply with its behavior in coherence evaluation, where it demonstrates a 74.6\% down-evaluation tendency. Other LLMs like Sonnet and Haiku show more moderate patterns, with balanced ratios between over and down-evaluation. CollabEval maintains the most balanced evaluation pattern, demonstrating its ability to mitigate extreme evaluation biases through collaborative assessment.

\subsubsection{Discussion about pair-wise comparison evaluation.}
Table~\ref{tab:pairwise} shows the comparison results of multi-agents and single LLM-as-a-Judge for pair-wise evaluation on two Arena datasets.  
This table employs four key metrics to assess model performance. \textbf{Accuracy} measures the percentage of correct predictions compared to ground truth. \textbf{Average Rounds} indicates the number of discussion iterations needed for consensus. \textbf{GT\_Win\_Pred\_Tie Ratio} represents the percentage of samples where ground truth indicates a clear winner but the model predicts a tie among all misevaluated samples, while \textbf{GT\_Tie\_Pred\_Win Ratio} shows the percentage of instances where ground truth indicates a tie but the model predicts a winner among all misevaluated samples. 

Our experimental results in Table~\ref{tab:pairwise} demonstrate CollabEval's superior performance across both datasets. On the Chatbot Arena Data, CollabEval achieves the highest accuracy of 60.2\% with 1.542 average rounds, significantly outperforming both Round-Table Eval (57.7\%) and single-LLM approaches. CollabEval shows balanced evaluation capabilities with a GT\_Win\_Pred\_Tie ratio of 50.00\% and a notably low GT\_Tie\_Pred\_Win ratio of 2.63\%, indicating its discrimination ability in ambiguous cases. For the Arena Human Preference Data, which presents more challenging evaluations, CollabEval maintains its performance advantage with 51.5\% accuracy and 1.517 average rounds, compared to Round-Table Eval's 48.7\% and single-LLM approaches ranging from 48.4\% to 50.5\%. While Single-LLM Llama3 70b shows competitive accuracy rates, its extreme ratios (53.85\%/0.00\% for Arena Data and 55.47\%/0.39\% for Preference Data) suggest potential bias in decision-making. Single-LLM Sonnet demonstrates more balanced performance but with lower accuracy (48.4\%) and higher GT\_Tie\_Pred\_Win ratio (13.95\%), indicating a tendency to make definitive judgments in ambiguous cases. These results consistently demonstrate that CollabEval's multi-agent approach, despite requiring additional computational rounds, provides more reliable and balanced evaluations compared to both round-table and single-LLM evaluation methods.
\subsection{Findings and Analysis}
\begin{figure}[t]
 	\centering
 	\includegraphics[width=0.48\textwidth]{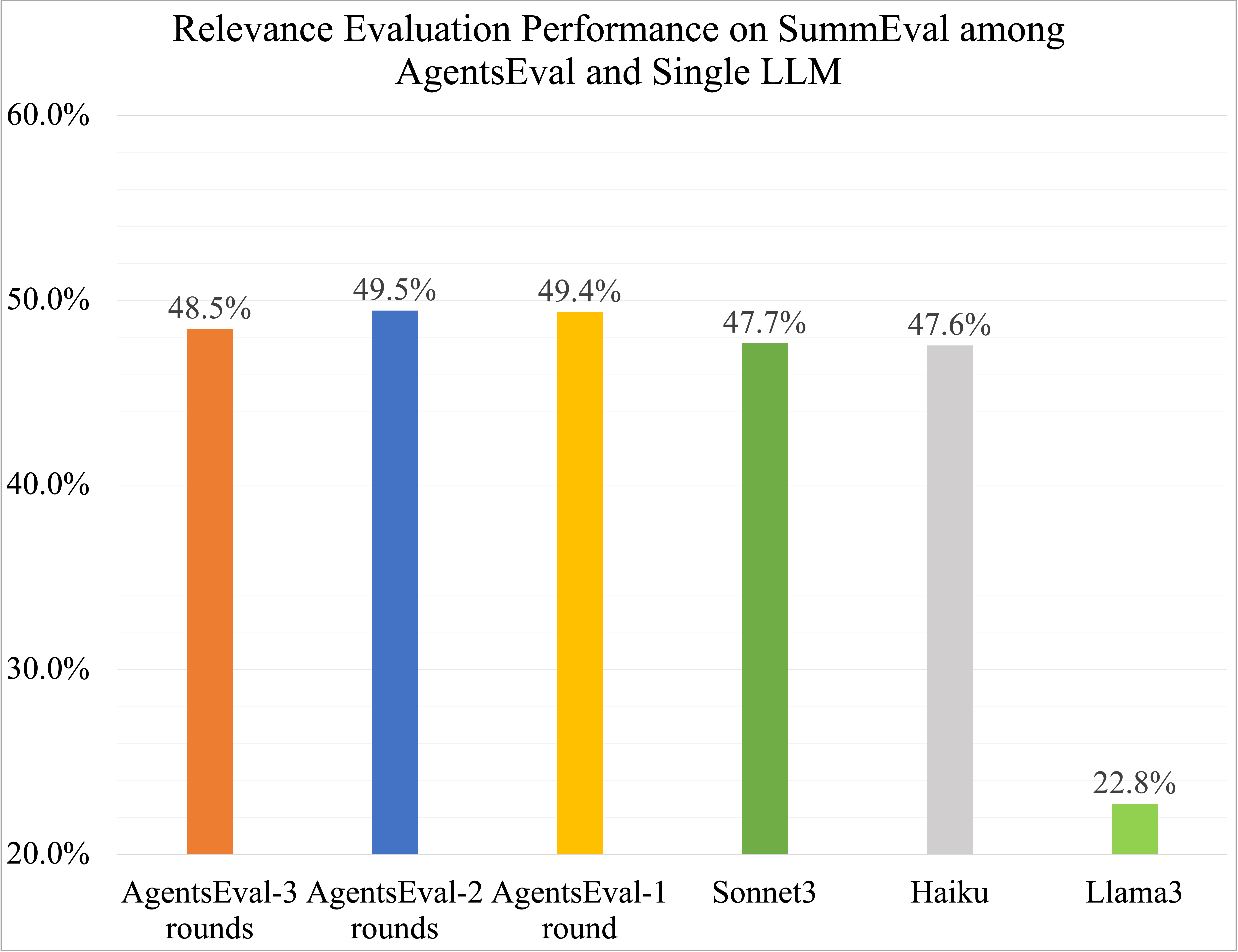}
         	\caption{Accuracy performance Analysis on relevance evaluation.
        } 
 	\label{fig:rounds}
 	\vspace{-0.2in}
 \end{figure}

 \begin{figure}[t]
 	\centering
 	\includegraphics[width=0.48\textwidth]{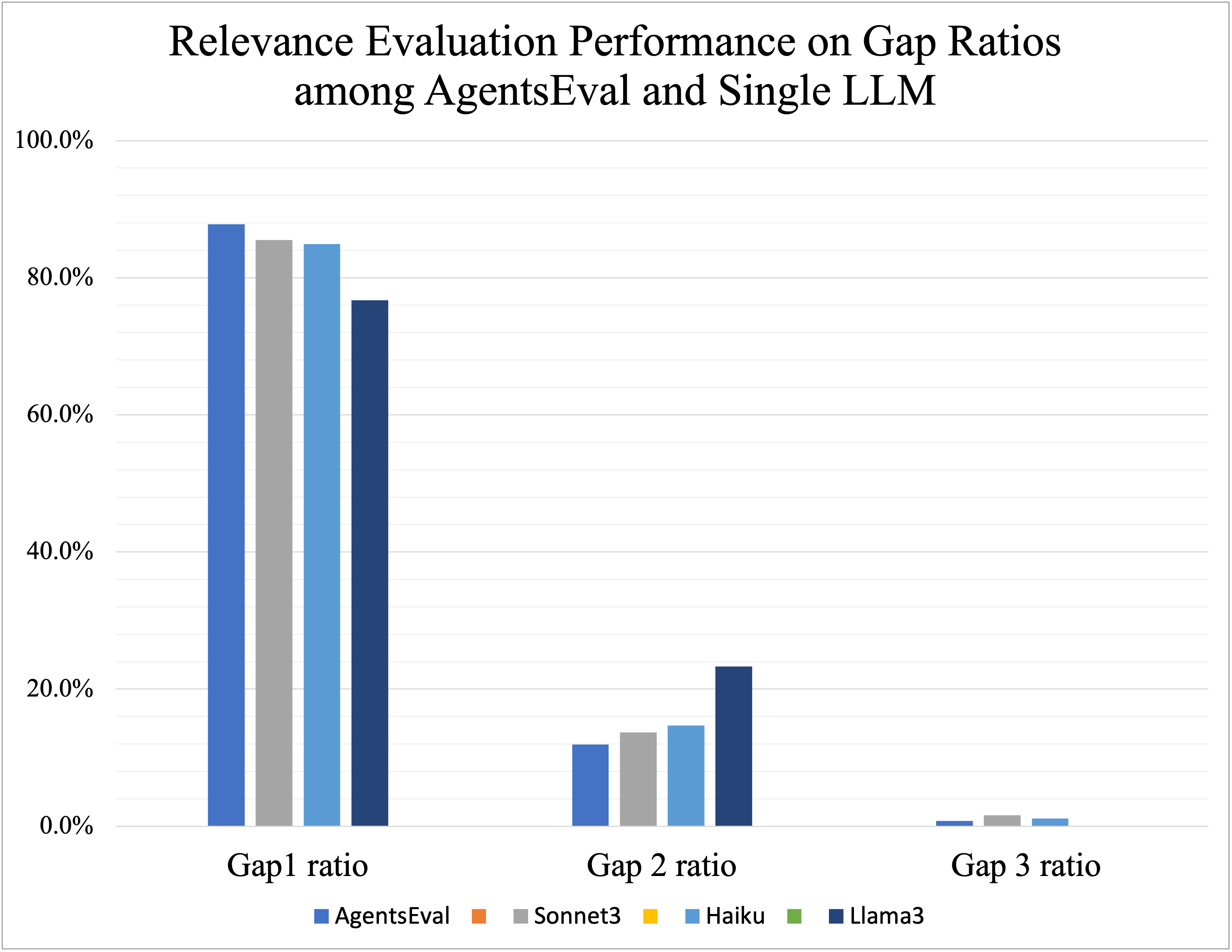}
         	\caption{Gap ratio performance analysis on relevance evaluation.
        } 
 	\label{fig:gaps}
 	\vspace{-0.2in}
 \end{figure}

  \begin{figure}[t]
 	\centering
 	\includegraphics[width=0.48\textwidth]{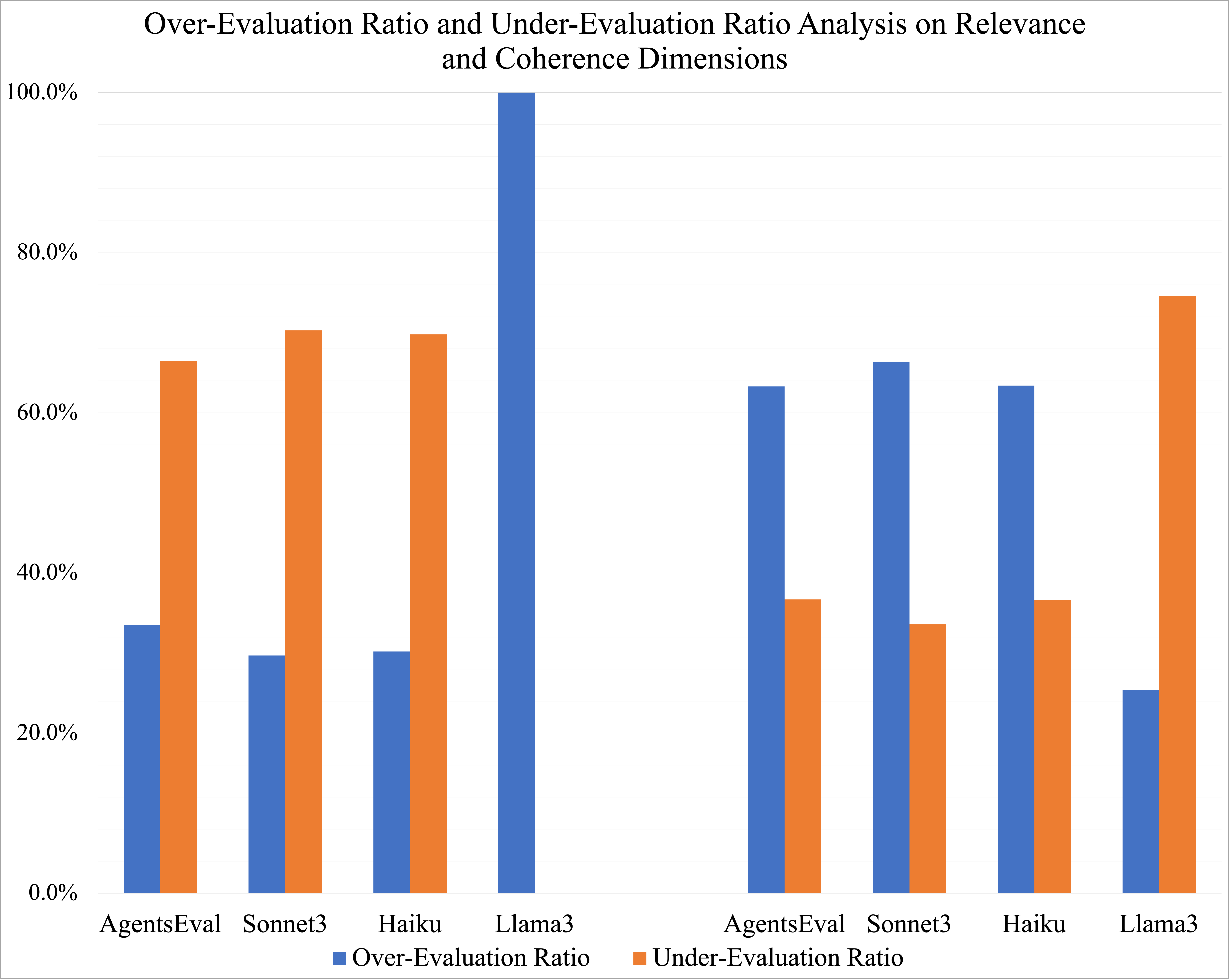}
         	\caption{Evaluation trends analysis on relevance (left) and coherence (right) evaluations.
        } 
 	\label{fig:trends}
 	\vspace{-0.1in}
 \end{figure}
\textbf{Discussion Rounds.}
The impact of discussion rounds on CollabEval's performance reveals several key patterns and underlying factors. In relevance evaluation, as illustrated in Figure~\ref{fig:rounds}, we observe a progressive improvement from one to three rounds: CollabEval with 1 round achieves 49.4\% accuracy, increasing to 49.5\% with 2 rounds, and slightly decreasing to 48.5\% with 3 rounds. This pattern demonstrates the trade-off between efficiency and accuracy, where initial collaboration brings significant improvements but faces diminishing returns beyond two rounds.

The diminishing returns phenomenon can be attributed to several key mechanisms. First, information saturation occurs as evaluators exchange most critical insights during early rounds, with subsequent rounds adding minimal new perspectives. This is evidenced by the Gap 1 ratio analysis in Table~\ref{tab:criteria}, where CollabEval achieves 87.8\% compared to single-model performances (Sonnet: 85.5\%, Haiku: 84.9\%), showing that major evaluation refinements happen early. Second, when compared to single-LLM performances (Sonnet: 47.7\%, Haiku: 47.6\%), even CollabEval with a single discussion round (49.4\%) outperforms these baselines, indicating that the multi-agent framework's primary benefits emerge from initial evaluation and first-round discussion.

These findings have important implications for practical deployment: while additional rounds of discussion can refine evaluations, the optimal configuration should balance the performance with reasonable computational overhead. This insight aligns with CollabEval's design principle of being cost-effective and efficient while maintaining comprehensive evaluation capabilities across various dimensions.
\\
\\
\noindent\textbf{Gap Ratio Analysis.}
The Gap Ratio analysis on relevance evaluation in Figure~\ref{fig:gaps} reveals significant patterns in evaluation precision across different models. CollabEval demonstrates superior performance with the highest Gap 1 ratio, followed by Sonnet, Haiku, and Llama3. This distribution pattern indicates several key findings about evaluation behavior. First, the close clustering of Gap 1 ratios among CollabEval, Sonnet, and Haiku suggests a consistent level of precision among advanced models, while Llama3's lower performance indicates a gap in relevance evaluation.

The progression of error severity provides further insights into model reliability. CollabEval shows a steep decline from Gap 1 to Gap 2 to Gap 3, indicating that when errors occur, they tend to be minor. This contrasts with Llama3's flatter distribution, suggesting less discrimination in error magnitude. The minimal occurrence of Gap 3 errors and Gap 4 errors across all models indicates that severe misjudgments are rare, though CollabEval maintains the lowest rate of such errors.

These findings suggest that while all models generally avoid severe misjudgments, CollabEval's collaborative approach leads to more refined evaluations with a higher concentration of minimal errors, demonstrating the effectiveness of multi-agent evaluation in maintaining precision while minimizing severe evaluation mistakes.
\\
\\
\noindent\textbf{Evaluation Patterns.}
The analysis of evaluation patterns in Figure~\ref{fig:trends} reveals distinct dimensional behaviors across different models. In the relevance dimension, CollabEval, Sonnet, and Haiku demonstrate a consistent tendency toward down-evaluation, with under-evaluation ratios of approximately 68.1\%, 70.3\%, and 69.8\% respectively in Table~\ref{tab:criteria}. This conservative evaluation approach suggests these models are more stringent in assessing relevance. Conversely, in the coherence dimension, these same models show a pronounced shift toward over-evaluation, with CollabEval showing a 63.3\% over-evaluation ratio, Sonnet at 66.4\%, and Haiku at 63.4\%, indicating a more lenient assessment of coherence qualities. 

Llama3 presents a particularly interesting case with extreme evaluation patterns that deviate significantly from other models. In relevance assessment, it shows a stark 100\% over-evaluation ratio, suggesting a consistent bias toward higher scores. This contrasts sharply with its coherence evaluation, where it demonstrates a 74.6\% under-evaluation tendency. These extreme patterns highlight two critical insights: first, the potential for individual models to develop strong biases in specific dimensions, and second, the importance of employing a balanced multi-agent approach to mitigate such extreme tendencies. 

The contrasting evaluation patterns between dimensions and models underscore the value of CollabEval's collaborative approach, which helps balance these inherent biases through multi-agent consensus, resulting in more nuanced and reliable evaluations across different dimensions.
\begin{figure}[t]
 	\centering
 	\includegraphics[width=0.49\textwidth]{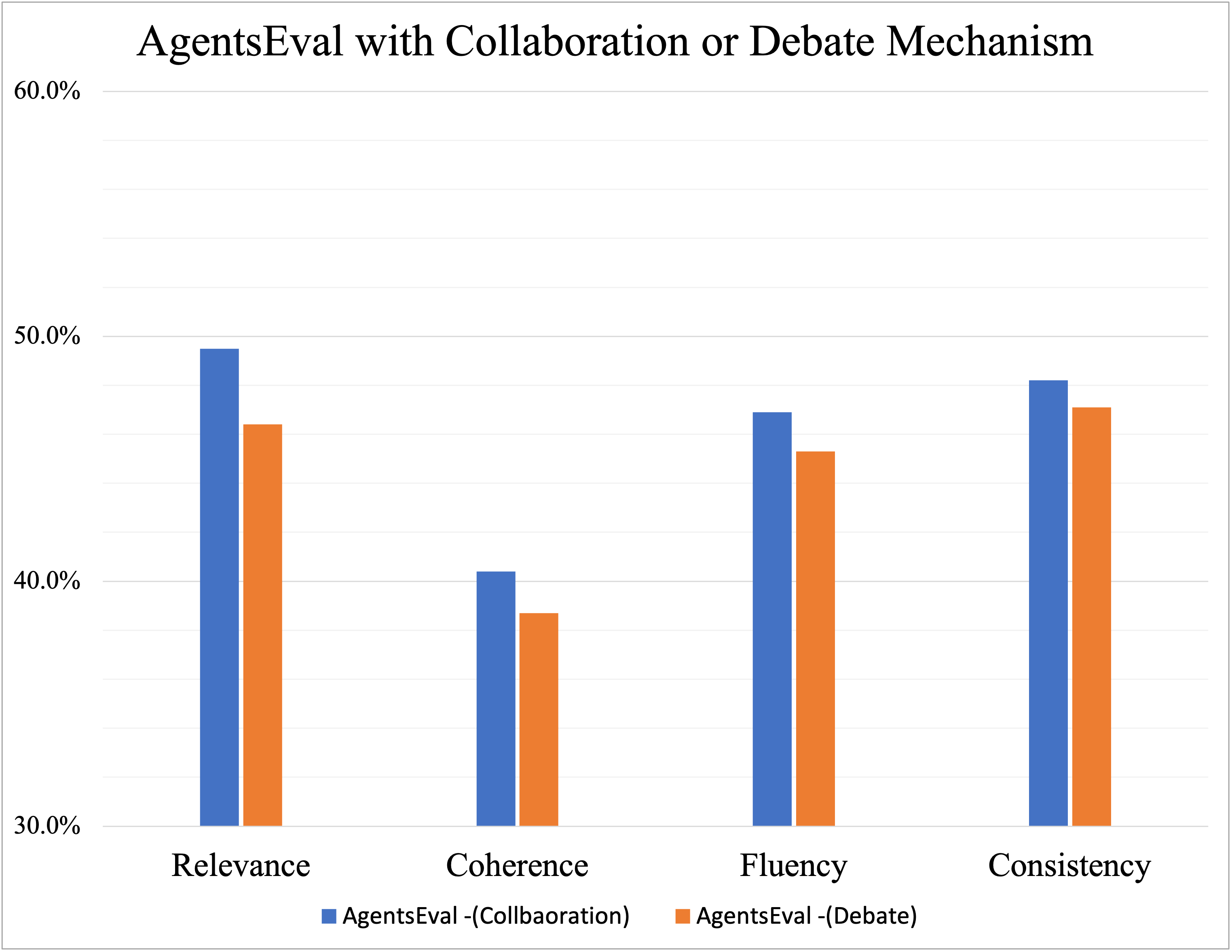}
         	\caption{Accuracy analysis of CollabEval with collaboration or debate mechanisms on SummEval dataset.
        } 
 	\label{fig:mode}
 	\vspace{-0.1in}
 \end{figure}

\noindent\textbf{Robustness and  Consistency.}
CollabEval further demonstrates remarkable robustness across evaluation scenarios, particularly evident in its ability to maintain consistent performance despite individual model limitations. In relevance evaluation, while Llama3 shows significant performance degradation (22.8\%) in Figure~\ref{fig:rounds} and Table~\ref{tab:criteria}, CollabEval maintains a strong accuracy of 49.4\% even with just one round of discussion. This performance stability extends across different dimensions, with CollabEval achieving 40.2\% in coherence evaluation compared to Llama3's 29.5\%, and similar patterns in other dimensions. 

The assembling mechanism operates through several channels. First, when one model shows extreme evaluation patterns (such as Llama3's 100\% over-evaluation tendency in relevance), CollabEval's collaborative framework effectively balances this through input from other evaluators, resulting in more moderate and accurate assessments (31.9\% over-evaluation ratio). Second, the multi-agent setup allows for cross-validation of evaluations, where stronger models can help correct the biases of weaker ones. This is particularly evident in the Gap ratio analysis, where CollabEval maintains the highest Gap 1 ratio (87.8\%) despite incorporating inputs from models with varying individual performances. 

These findings suggest that CollabEval's robust performance is not merely an averaging effect but rather an orchestration mechanism that leverages the strengths of each model while mitigating their individual weaknesses through collaborative evaluation.
\\
\\
\noindent\textbf{Collaborative Advantage.}
We last explored the mechanisms of CollabEval, collaboration mechanism or debate mechanism, as shown in Figure~\ref{fig:mode}. The experimental results demonstrate that CollabEval's collaborative approach consistently outperforms the debate mechanism across all evaluation dimensions. In relevance assessment, the collaborative mechanism shows a clear advantage in this critical dimension. Similar patterns emerge in coherence, fluency, and consistency evaluations. 

This consistent performance gap suggests that collaboration, where agents focus on sharing insights and building upon each other's evaluations, is more effective than competitive debate mechanisms. The collaborative approach's superior performance can be attributed to its emphasis on constructive information sharing and consensus building, rather than adversarial discussion. This aligns with our findings from the criteria-based evaluation results, where CollabEval's collaborative framework demonstrates robust performance across different dimensions while maintaining reasonable computational efficiency through optimal round limitation.

\section{Conclusion}

In this paper, we propose CollabEval, a novel multi-agent framework for evaluating AI-generated content. Through extensive experiments, we demonstrate that CollabEval consistently outperforms single-LLM approaches across multiple dimensions, achieving optimal performance with several discussion rounds and showing superior capability. The framework's robust performance, even when individual models struggle, validates the effectiveness of our collaborative evaluation approach. Future work could explore extending the framework to more complex evaluation scenarios and investigating the impact of different model combinations on evaluation outcomes.

\clearpage
\newpage
\bibliography{aaai25}

\clearpage 
\newpage 

\clearpage
\newpage
\appendix

\end{document}